# Integrative Use of Computer Vision and Unmanned Aircraft Technologies in Public Inspection:

Foreign Object Debris Image Collection

Travis J. E. Munyer
Department of Computer Science, University of Nebraska Omaha
tmunyer@unomaha.edu

Daniel Brinkman
Department of Computer Science, University of Nebraska Omaha
dbrinkman@unomaha.edu

Chenyu Huang
Aviation Institute, University of Nebraska Omaha
chenyuhuang@unomaha.edu

Xin Zhong
Department of Computer Science, University of Nebraska Omaha
xzhong@unomaha.edu

## ABSTRACT

Unmanned Aircraft Systems (UAS) have become an important resource for public service providers and smart cities. The purpose of this study is to expand this research area by integrating computer vision and UAS technology to automate public inspection. As an initial case study for this work, a dataset of common foreign object debris (FOD) is developed to assess the potential of light-weight automated detection. This paper presents the rationale and creation of this dataset. Future iterations of our work will include further technical details analyzing experimental implementation. At a local airport, UAS and portable cameras are used to collect the data contained in the initial version of this dataset. After collecting these videos of FOD, they were split into individual frames and stored as several thousand images. These frames are then annotated following standard computer vision format and stored in a folder-structure that reflects our creation method. The dataset annotations are validated using a custom tool that could be abstracted to fit future applications. Initial detection models were successfully created using the famous You Only Look Once algorithm, which indicates the practicality of the proposed data. Finally, several potential scenarios that could utilize either this dataset or similar methods for other public service are presented.

## 1 INTRODUCTION

Integrating state-of-the-art advances in technology has provided consistent benefit towards major city operations. Artificial intelligence is one of the major incorporating areas for smart city infrastructures. Smart city concepts include various systems such as New York City's Midtown in Motion, a traffic monitoring system that has helped to reduce traffic congestion [12]. New York City has also implemented a water usage monitoring system that has over 800,000 gauges in which the city's water consumption data is measured and aggregated. In the future, currently trivial security camera systems may be able to detect crimes like assault and proceed to track and monitor suspects.

Computer vision, a scientific field studying high-level understanding from digital images or videos, is behind many recent advances in artificial intelligence. Together with machine (deep) learning [23], object localization (*i.e.*, identifying the locations of objects in an image) and object detection (*i.e.*, detecting instances of semantic objects of a certain class in images) has attracted increasing attention and has demonstrated its supreme performance in many application scenarios [9, 11]. Developing object detection or localization models is an important first step in many inspection systems for smart cities [22]. However, building these models often require massive, annotated data and thus there is a crucial demand for computer vision databases in this research community, *e.g.*, Foreign Object Debris (FOD) datasets [19].

Unmanned aircraft systems (UAS), commonly referred to as drones, are any aerial vehicles that do not contain an onboard pilot. Without an onboard pilot, UAS can be significantly smaller than manned aircraft, fly in more dangerous situations, and carry a variety of different payloads. UAS technology is an efficient and novel way to collect massive quantities of image data because of its capabilities. UAS are usually controlled by a remote pilot manually or autonomously using flight control software and Global Positioning System (GPS). These techniques usually command the UAS to fly to a specific GPS coordinate or along a path of coordinates to a destination. Therefore, most automatic UAS navigation systems are limited to spaces covered by GPS service. Other navigation methods can involve preprogrammed movements, following a painted line based on computer vision, inertial navigation, or other techniques. With the significant miniaturization of advanced integrated electronics, rapid development of power storage and materials, and continuously declining equipment cost, small UAS are becoming more common and promising in civilian use.

Research in FOD detection or localization using computer vision methods is becoming common [25–27]. However, there is not a publicly released dataset that is sufficient for complex research applications. Researchers have been repeatedly creating private FOD datasets for individual algorithm exploration. Without a common dataset, it is challenging to compare and evaluate the performance of new algorithms. We propose to solve this issue by initializing a common dataset of FOD. In the literature, although there is one existing FOD dataset, it mainly concentrates on the specialized purpose of material recognition [24]. It can be too small (about 3,500 images) to support various object detection tasks in machine learning. Moreover, existing datasets (e.g., 3 classes in [24]) did not cover a variety range of categories of FOD. In contrast, the initial collection of our FOD dataset contains 11 object classes and is expanding as we continue to incorporate all categories of FOD defined by the Federal Aviation Administration (FAA) [3].

In the proposed public inspection, one major purpose is to train a computer vision model that detects objects on a runway or taxiway, which requires sizable training examples to teach the model target object patterns. To this end, the dataset needs to be extensible, meaning that adding objects to existing datasets must be significantly simple. Also, to provide benefit of categorization, the computer will learn per-object as opposed to learning all objects at once. Typically, machine learning processes of object detection consists of inputting a dataset of images with the location of each object given as coordinates that form a bounding box. Each bounding box must be manually drawn on every image to create this dataset. After studying these images, the model can infer the location of bounding boxes on new images without the input bounding box. If the model estimates an incorrect bounding box, it will reassess the estimation it made initially and attempt another approximation after studying the annotated images further. Over time the model will learn to understand which output coordinates are correct, which are incorrect, and will gradually become increasingly proficient at making estimations.

Besides the most apparent large data scale obstacle, the format of the data itself can present problems if left unoptimized. A universal format would certainly need to be standardized, as it would drastically reduce overhead caused by converting between formats and increase the scalability of the data. Presumably, cities would also need to communicate between each other as well. This would necessitate reliable networks with the ability to persistently handle large datasets. The final major setback in implementation of smart city technology is the cost. The Midtown in Motion project alone accrued nearly $3 million in fees and will likely cost millions more [12]. One of the goals of this research is to create an easy-to-follow framework which minimizes data usage to reduce requirements on networks. Utilizing computer vision technology that can be implemented on more economical hardware would also be a major benefit towards mass adoption.

## 2 APPLICATIONS IN PUBLIC SERVICES

This section introduces application examples of integrative use of computer vision and UAS technology focusing on FOD detection and other potential applications in public services.

### 2.1 Potential Applications in Public Service

The presented strategy of UAS based computer vision could be applied to a variety of critical public services with necessary datasets for technology preparation. Bridge inspection, as one example, is a crucial practice to prevent bridge collapse and similar dangerous events. In current practice, inspection personnel must search for defects in all sections of the bridge when conducting the inspection. Examining the underside of a bridge can be unsafe and expensive. Leveraging the advantages of mobility and versatility of UAS technology, UAS could be employed to inspect the elusive sections of bridges to prevent unnecessary risk to personnel [20]. Furthermore, incorporating computer vision into UAS platform expects to further enhance the technological capability and efficiency of UAS in bridge inspection by preventing human errors and overlooking potential bridge damages. While human examination would still be necessary at times, regular inspections could be automated using precision detection and navigation techniques embedded in UAS. Any automated inspections using these methods could capture complete video data for review by authorized personnel.

Applying these methods to traffic monitoring could also prove a beneficial practice. Standalone cameras already exist in many existing roadways, and computer vision enabled UAS could provide a mobile inspection service. Computer vision, if applied to these sensors, could provide detailed information about roadway usage and peak traffic hours [7]. Utilizing UAS to monitor traffic would allow more roads to become monitorable, as aerial vehicles would not be limited to a single location. Traffic intersections could be monitored for debris to expedite its removal or to alert emergency personnel of collisions [21]. Video motion analysis could be included to estimate the velocity of vehicles, which would provide a cheaper alternative to manual speed limit monitoring. The data collected from UAS and stationary cameras would provide detailed reports and mappings about traffic that could prove crucial in road development and other urban planning situations.

Cities are becoming overwhelmingly large, which contributes to the growing complexity of planning urban development. Currently, personnel are deployed to perform manual evaluation of the city environments for urban planners, which can become costly and tedious. A study discusses a particularly interesting machine learning method that utilizes information such as maintenance issues (cracks), building materials, and industrial precision and craftsmanship to rate the beauty of an urban area [8]. However, the data they use comes from Baidu Maps (Similar to Google Maps) which may not get updated frequently [8]. If their models were modified to perform real-time analysis from the viewpoint of UAS, these methods could become a reliable and consistent resource in planning expanding metropolises.

## 2.2 Foreign Object Debris

According to the FAA, FOD is defined as, "Any object, live or not, located in an inappropriate location in the airport environment that has the capacity to injure airport or air carrier personnel and damage aircraft" [3]. The FAA's report discussing FOD detection equipment lists typical types of FOD including:

- "aircraft and engine fasteners (nuts, bolts, washers, safety wire, etc.);
- aircraft parts (fuel caps, landing gear fragments, oil sticks, metal sheets, trapdoors, and tire fragments);
- mechanics' tools;
- catering supplies;
- flight line items (nails, personnel badges, pens, pencils, luggage tags, soda cans, etc.);
- apron items (paper and plastic debris from catering and freight pallets, luggage parts, and debris from ramp equipment);
- runway and taxiway materials (concrete and asphalt chunks, rubber joint materials, and paint chips);
- construction debris (pieces of wood, stones, fasteners, and miscellaneous metal objects);
- plastic and/or polyethylene materials;
- natural materials (plant fragments and wildlife); and
- contaminants from winter conditions (snow, ice)." [2]

This list of examples is not all inclusive, as FOD can be any object of any material, size, or color [3]. Causes of FOD are varied, but can be created by weather, personnel, the environment (wildlife, snow, ice), any equipment operating on the airfield (aircraft, maintenance equipment, construction equipment, etc.), or existing aircraft infrastructure (cracked pavement, broken runway lights, etc.) [2].

*2.2.1 Risk to Airport Operations.* Boeing reported an estimated $4 billion in damages per year, including damage to engines caused by ingesting debris and/or the cost of removing aircraft from service for repairs [4]. FOD can also destroy aircraft tires or become lodged in other locations of the aircraft and potentially prevent proper function [4]. Aircraft engines are powerful enough to launch small pieces of FOD. When FOD is launched, it can cause damage to surrounding buildings and aircraft or injure local personnel [3]. FOD generally causes minor damage to aircraft blades or tires, but in extreme cases FOD can cause major accidents, usually when landing or taking off. The FAA mentions an incident where major loss of life occurred:

> "On July 25, 2000, an Air France Flight 4590 departing Charles de Gaulle International Airport ran over a piece of titanium debris from a Continental DC-10, shredding a tire and slamming rubber debris into the plane's fuel tank. The subsequent leak and fire caused the Concorde to crash, killing 100 passengers, nine crewmembers, and four people on the ground." [4]

Proper FOD management is essential to preventing injury or death to personnel and passengers. Preventing FOD from littering airfields can reduce cost accrued from damage to aircraft engines or fuselages. Proper FOD prevention and removal is also required by law for airports that serve aircraft designed for more than 9 seats [5]. Daily FOD inspection and prompt removal is necessary for these airports to be awarded required certifications [5].

*2.2.2 Potential Computer Vision Applications.* Current FOD detection systems only recognize the existence of debris in targeted areas without automatically labeling detected objects. Utilizing computer vision, automated recognition could be expanded to personalized reports based on the object detected. Once FOD is detected by stationary cameras or by small UAS with mounted cameras, a report could be generated with a classification of the object. This could be expanded to include estimated dimensions of debris, as well as other calculations and details deemed useful. One such calculation would be an estimated mass measurement. Applications already exist that can measure objects using a camera, so volume could be calculated based on these estimations. Since mass is density times volume, a density measurement also is necessary. This missing value could be obtained by linking an average density to each FOD category, calculated by explicitly weighing objects over time and recursively updating this average in a database. Of course, this method could only provide estimations with an allowed margin of error due to the nature of the calculation.

Another potential application of computer vision for FOD detection would entail automating the removal of debris. Automating the detection and removal of FOD would reduce costs induced by aircraft damage and prevent injury or death to victims of FOD related accidents. To automate the removal of FOD using UAS or even unmanned ground vehicles (UGVs), an automated navigation system would first need to be implemented in the vehicles. Automated navigation is a complex subject with a few interesting technical papers discussing the technology. Majumdar et al. explore one potential theory for an automated guidance system, which leverages another form of machine learning. Majumdar et al. utilizes a similar framework for both the navigation of a UAS, and the task of grasping an object [10]. One issue with the grasping framework presented in the paper can be found in the mass measurement used for crucial grasping calculations [10]. The mass used in their calculations is a randomly generated number in the range [0.05, 0.15] kg [10], which could cause inaccurate results in a real setting. This problem could be solved by using the suggested estimated mass calculation presented to achieve a more tailored result. A UAS could also use existing methods for navigation such as following painted lines, but a more universal approach to guidance would be beneficial for an inspection system. Once the navigation and grasping system is developed, FOD could be automatically detected and removed from airport grounds by utilizing computer vision enabled UAS.

*2.2.3 Challenges behind implementation.* Before implementing solutions for these issues, some problems could arise from implementing UAS as automated FOD removal and detection units. Once UAS are patrolling airspace shared with manned aircraft, measures would need to be taken to prevent these UAS from interfering with existing traffic, and/or becoming FOD themselves. A long-term solution would involve developing a robust software application that either performs actions based on inputted air traffic schedules and/or involves the UAS recognizing the aircraft themselves.

The UAS could then move out of designated paths, or to specified storage areas. Utilizing only the air traffic schedules could cause issues also. If any irregularities in the schedule occur, the UAS could become FOD without other preventative measures.

UAS would be negatively affected by weather conditions such as snow, rain, and wind. Snow and rain could conceal FOD, irregulate existing UAS flight patterns, and cause other unforeseen issues. Wind could displace light FOD and cause major issues to UAS flight stability depending on the size and weight of the vehicles.

If automated FOD removal was implemented using these methods, objects of excessive weight would still have to be removed manually. Staff on FOD teams would still be necessary for this purpose but could also be assigned other tasks. Objects of excessive weight may cause issues with this theory, but it seems the majority of FOD is relatively small in normal circumstances.

A combination of machine learning for navigation and general FOD recognition tasks could prove quite computationally expensive, especially if run directly on the UAS local hardware. This could be solved by determining results on remote machines. However, this would require a wireless network that covers an airfield, with enough speed to determine results in real-time.

## 3 FOREIGN OBJECT DEBRIS DETECTION CASE STUDY

In these subsections, we review existing FOD literature, such as existing FOD detection equipment, and examine the results of our dataset.

### 3.1 Existing FOD Detection Equipment

Four manufacturers have developed automated FOD detection equipment [4], including XSight's FODetect based in Boston, MA; Trex Enterprises' FOD Finder based in San Diego, CA; QinetiQ's Tarsier Radar based in the United Kingdom; and Stratech's iFerret system based in Singapore. These cover the major technology types tested by the FAA and the University of Illinois Center of Excellence for Airport Technology (CEAT), namely millimeter wavelength radar, high-definition video, hybrid video-radar, and mobile FOD detection technology [13–16]. The FAA defines minimum performance requirements for each of these automated FOD detection technologies in their Advisory Circular (AC) 150/5220-24 [2]. A requirement specified by the FAA mentions that detection systems must inspect surfaces between aircraft movements, which is a time specified by individual airports [2]. A common detection time given by airports is generally within 4 minutes of the placement of FOD [2].

FODetect uses combined video and radar technology to detect FOD [4]. This technology is mounted next to runway lights at the edges of the runway [4]. When FODetect locates FOD, it sends both an alarm message and a video image to the operator [4]. During performance trials, FODetect had an average detection time of 35 seconds [16]. The average difference between the reported location of FOD and the actual location was 5.02 ft (1.53 m), and the greatest difference between the reported location and the actual location of FOD was 25.6 ft during trials (7.80 m) [16]. The required location accuracy must be within 16 ft (5.0 m) of the actual FOD location [16]. FODetect performed as expected during rainfall and snowfall during CEAT and FAA testing after snow was cleared when applicable [16]. FODetect is about the size of runway lights, which does not excessively interfere with airport safety zones [16].

FOD Finder uses radar technology to locate FOD. FOD Finder is a mobile FOD detection system and is usually vehicle mounted. Once FOD is detected, FOD Finder will send audible and visual alerts to the operator [13]. The system has an optional vacuum that can retrieve and catalogue items, or the operator can opt to remove the item manually [13]. The detection scan is directed at the front of the vehicle in an 80-degree wide cone shape that is about 650 ft (200 m) long [13]. FOD Finder performed as expected during and after rainfall and snowfall events during testing [13]. Since FOD Finder is vehicle mounted, it is not capable of 24/7 surveillance like the fixed solutions. The equipment also needs to be operated by a driver. Because it is vehicle mounted, a single equipped vehicle can be directed at different locations around the airport.

Tarsier Radar uses millimeter wavelength technology to scan pavement for FOD [4]. This unit sends an alarm message with the location of the object after FOD is detected [4]. Tarsier is mounted on large towers (tower height can vary per airport), which can interfere with airport safety zones if too tall [14]. If the towers are not placed high enough, coverage could be lost in pavement locations that are not completely flat [14]. These units have a possible range of 1.25 miles (2 km) but are rated to 0.6 miles (1 km) for accurate performance [14]. Tarsier had issues consistently detecting FOD in heavy rain or when snow and ice accumulated during CEAT & FAA testing [14]. FOD detection occurred within a scan time of 1 minute and detections were confirmed in less than 4 minutes [14]. Tarsier exceeded FAA requirements of location accuracy within 16 feet (5.0 m), with an average difference between the FOD's actual location and the reported location of the FOD being 3 ft (0.91 m) [14]. The greatest difference between the FOD's actual location and the reported location was 7 ft (2.13 m) during testing [14].

IFerret uses video technology alone to detect FOD on surfaces [4]. IFerret is mounted on towers up to 500 ft (175 m) away from the target surface, and each sensor is designed to cover approximately 1100 ft (330 m) of the target surface [15]. IFerret can work on runways, taxiways, aprons, and in ramp areas [15]. During testing, iFerret performed as expected in all lighting conditions and in snow, rain, and ice after snow/ice removal when applicable [15]. When iFerret detects FOD, the operator receives audible and visual alerts, with a provided location and image [15]. IFerret had an average location accuracy of approximately 0.82 ft (0.25 m), which exceeded the FAA requirement of 16 ft (5.0 m) [15]. The greatest difference between reported location and actual location during testing was 1.56 ft (0.48 m). Average scan time was about 70 seconds, within the common airport-recommended time of 4 minutes [15].

Each system shares an inability to automatically classify detected FOD. Also, each device does not provide a completely accurate GPS location, which is expected to a degree. The only device with a system put in place that semi-automatically removes FOD is FOD Finder with its vacuum option. The other devices do not include a system that automatically removes FOD, which could be greatly beneficial if developed.

Most airports today have not yet implemented automated FOD detection due to its high cost. A scalable solution that requires less installation time and is cheaper than existing methods could be a solution to this issue. Compared to existing detection equipment, UAS-based detection systems could prove cheaper to install and produce. Manual methods of FOD detection rely on human performance. According to the Galaxy Scientific Corporation (GOC) and the Office of Aviation Medicine (OAM), the performance of manual methods can be affected by human mistakes and failures [6]. This can include introducing FOD during maintenance, committing unsafe acts that can have consequences such as introducing FOD to the environment, and failures that can occur due to simply missing details, which can often be due to exhaustion [6]. Using only manual methods of FOD detection is risky since personnel naturally make mistakes.

### 3.2 Methodology

To create a dataset that can be used to build a low-cost, automated, and efficient FOD detection system based on computer vision and machine learning, data of common FOD objects were collected using both a portable camera and a UAS which shot videos from multiple angles of the objects. These images were collected in a video format at Millard Airport (KMLE) in Omaha, Nebraska. Computer Vision Annotation Tool (CVAT) [17] was used to manually annotate bounding box labels for the locations of objects in the videos. CVAT also allowed us to export these bounding box annotations into easily usable formats. Data was exported originally in standard (*i.e.*, Pascal VOC) format. After developing and using a tool to separate the videos into individual images, we were left with a dataset of 14,260 object instances, each having an accompanying frame detailing the annotated labels including the location and class.

A video has several images contained in each second, commonly known as frames-per-second (FPS), that combine to produce a motion display. For example, if we input a 30.6 second video with an FPS of 30, our tool would output a total of 918 images. This forms the framework for creating an image dataset using video data. We resize the images and annotations to 400x400 resolution to facilitate a unified-size modeling, while the original images are also made available. Our resizing tool also allows us to re-validate the accuracy of modified dataset annotations. The resize tool can optionally display all the images with their bounding boxes. This simply allows images to be visually inspected as the dataset is resized. The annotations were then split into separate training and validation datasets. We split the dataset into 75 percent training and 25 percent validation. This allowed for a training set of approximately 10,000 images spread across our different classifications of objects. Table 1 shows the total counts of all objects in the FOD dataset.

The images were split at random into the respective training and validation groups, although they were kept within their class identifications to ensure the 75/25 split occurred for each object type.

### 3.3 Results

To illustrate our dataset's practicality, we have tested a widely used object detection algorithm in computer vision and deep learning (YOLOv5 [18]) on all created data. The model was created using the "medium" version of YOLOv5, or YOLOv5m to obtain accurate results while outputting a model that could run quickly on affordable hardware. We trained this model for 100 epochs, which means the entire training dataset was iteratively applied 100 times to adjust the model parameters for a higher accuracy. The validation images are only used to confirm the model performance and hence they were not included during the training. This led to results of 95.2% precision, 99.3% recall, and a mean average precision of 99.1% on the validation dataset. Precision is measured as the percentage of correctly predicting an object, with recall being the percentage of all objects detected. Mean average precision, or mAP, is measured by computing the mean of the average precision for each object. This indicates that the final model infrequently had false positives, but more importantly, it accurately detects most of the objects.

**Table 1: Object Instances in FOD Dataset**

| Object | Count |
| --- | --- |
| Bolt | 3118 |
| Nut | 520 |
| Metal Sheet | 394 |
| Adjustable Clamp | 544 |
| Washer | 1797 |
| Luggage Tag | 1616 |
| Plier | 2099 |
| Nail | 1005 |
| Hose | 294 |
| Bolt/Nut Combination | 514 |
| Wrench | 2359 |
| Total | 14260 |

Successful tests of YOLOv5 indicate an important extensibility. That is, our proposed dataset can be used to study mainstream methods or even variants/newly proposed methods for FOD detection. Figure 1 presents sample FOD images and subsequent detection results. Remarkably, in some cases, the model accurately detects some objects while it is even hard to see them in human vision.

Balancing the number of instances in each collected object class, expanding the object categories according to FAA regulation to reflect the comprehensive nature of FOD, and including FOD image data in various and/or extreme conditions (such as raining) will be our next-step work.

While we will continue to expand and improve upon our dataset, an important result of our process is a publicly documented method in which video data can be split into images and used to expand a dataset following our format. Database size is important for accurate machine learning, and therefore the ability to have an open set of instructions with which to add more data means that our dataset can continue to be improved in a simple manner. These instructions can be found on our dataset's GitHub page. These instructions can fit any target application if input data is in the video format. The final model can be then deployed and used on easily attainable hardware, leading to a solution that is adaptable, scalable, and implementable.

## 4 CONCLUSION AND DISCUSSION

Several important public services are still performed manually, while technology capable of automating and enhancing these functions are nearly in reach. The combination of UAS and computer vision technology will be an integral resource for future smart cities and for automating several public services. Many smaller steps will be required to eventually achieve this sizable goal, such as the creation of this FOD dataset. During initial modeling procedures, we found that our dataset performed exceptionally with the YOLOv5 algorithm. This dataset and its resultant models will soon need to be employed in practical computer vision experiments that utilize UAS technology as an image capturing device. As a case study, we began applying computer vision and UAS concepts to FOD. To summarize, the major contributions of this paper are organized as follows:

- Introduced a new concept for FOD detection using computer vision and UAS techniques.
- Proposed a theoretical framework for future automated FOD management. The extensibility from FOD to other applications is also discussed.
- Initialized a new dataset of various objects that commonly become FOD in airports.

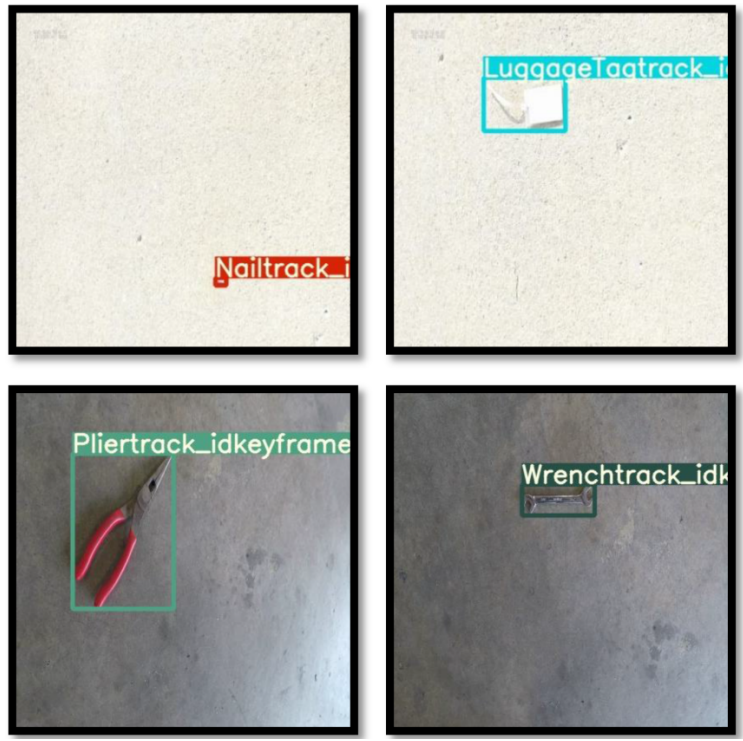

**Figure 1: Examples of output bounding boxes from the validation dataset during FOD model experimentation.**

A consistent data format would reduce both the computational and financial cost of machine learning implementation in future smart cities. This format should be expandable, as data required by smart cities may change. Our data format has consistency and expandability in mind, especially for video data. UAS that are designed to inspect transportation services or infrastructure should collect video data. This video data could be recycled and used to expand datasets, which means less personnel tasked with manual data collection. The practicality of data collected during real processes could prove naturally effective in updated detection algorithms. Implementing several datasets following a consistent format allows developers to utilize the same code and processes for several applications. Increasing the potential efficiency of developers reduces overall labor requirements, which is beneficial to implementation processes.

Datasets available to the public enable researchers to quickly design algorithmic experiments. Dataset creation can be labor intensive, and using a public dataset allows developers or researchers to skip unnecessary steps. Also, the results of new algorithms are easier to compare when using a public dataset. Datasets themselves can modify performance, so it is difficult to compare results without using a common dataset as a constant.

Deploying computer vision and UAS as an automated detection medium for FOD has a variety of potential applications that could facilitate airport safety and daily operations. As discussed in the potential applications section, UAS-based inspection could be abstracted to expand inspection/monitoring techniques designed previously. Furthermore, deepening this research path will open ways for more general development such as auto-removal of FOD after inspection. To progress towards these broader goals, existing processes and technology currently utilized by airports must be further analyzed. Continuing this research path could involve expanding areas such as automated UAS navigation and item retrieval, practical experiments using computer-vision-enhanced UAS to detect FOD, and the enhancement of general FOD detection algorithms.

# 5 RESOURCES

Our FOD Dataset GitHub page: https://github.com/FOD-UNOmaha/FOD-data

# ACKNOWLEDGMENTS

This work is funded by the NASA Nebraska Space Grant (Federal Award #NNX15AI09H).